\documentclass[10pt,twocolumn,letterpaper]{article}

\usepackage{iccv,times,graphicx,amsmath,amssymb,booktabs,tabulary,multirow,overpic,xcolor}
\definecolor{citecolor}{RGB}{34,139,34}
\usepackage[pagebackref=false,breaklinks=true,letterpaper=true,colorlinks, citecolor=citecolor,bookmarks=false]{hyperref}
\iccvfinalcopy

\usepackage[british,english,american]{babel}

\usepackage{listings}
\definecolor{codegreen}{rgb}{0,0.5,0}
\definecolor{codeblue}{rgb}{0.25,0.5,0.5}
\definecolor{codegray}{rgb}{0.6,0.6,0.6}

\newcommand{\app}{\raise.17ex\hbox{$\scriptstyle\sim$}}

\newcolumntype{x}[1]{>{\centering\arraybackslash}p{#1pt}}

\newlength\savewidth\newcommand\shline{\noalign{\global\savewidth\arrayrulewidth
  \global\arrayrulewidth 1pt}\hline\noalign{\global\arrayrulewidth\savewidth}}

\makeatletter\renewcommand\paragraph{\@startsection{paragraph}{4}{\z@}
  {.5em \@plus1ex \@minus.2ex}{-.5em}{\normalfont\normalsize\bfseries}}\makeatother

\usepackage{xspace}
\newcommand{\f}{$^\text{*}$\xspace}

\begin{document}

\title{Group Normalization}
\author{
 Yuxin Wu \qquad\qquad Kaiming He \vspace{3mm}\\
 Facebook AI Research (FAIR)
\\
{\tt\small \{yuxinwu,kaiminghe\}@fb.com}
\vspace{-1.5em}
}

\maketitle

\begin{abstract} \vspace{-.5em}
Batch Normalization (BN) is a milestone technique in the development of deep learning, enabling various networks to train. However, normalizing along the batch dimension introduces problems --- BN's error increases rapidly when the batch size becomes smaller, caused by inaccurate batch statistics estimation. This limits BN's usage for training larger models and transferring features to computer vision tasks including detection, segmentation, and video, which require small batches constrained by memory consumption. In this paper, we present Group Normalization (GN) as a simple alternative to BN. GN divides the channels into groups and computes within each group the mean and variance for normalization. GN's computation is independent of batch sizes, and its accuracy is stable in a wide range of batch sizes. On ResNet-50 trained in ImageNet, GN has 10.6\% lower error than its BN counterpart when using a batch size of 2; when using typical batch sizes, GN is comparably good with BN and outperforms other normalization variants. Moreover, GN can be naturally transferred from pre-training to fine-tuning. GN can outperform its BN-based counterparts for object detection and segmentation in COCO,\footnote{\url{https://github.com/facebookresearch/Detectron/blob/master/projects/GN}.} and for video classification in Kinetics, showing that GN can effectively replace the powerful BN in a variety of tasks. GN can be easily implemented by a few lines of code in modern libraries.
\end{abstract}

\vspace{-1em}
\section{Introduction}

Batch Normalization (Batch Norm or BN) \cite{Ioffe2015} has been established as a very effective component in deep learning, largely helping push the frontier in computer vision \cite{Szegedy2016a,He2016} and beyond \cite{Silver2017}. BN normalizes the features by the mean and variance computed within a (mini-)batch. This has been shown by many practices to ease optimization and enable very deep networks to converge. The stochastic uncertainty of the batch statistics also acts as a regularizer that can benefit generalization. BN has been a foundation of many state-of-the-art computer vision algorithms.

\begin{figure}[t]
\centering
\includegraphics[width=.92\linewidth]{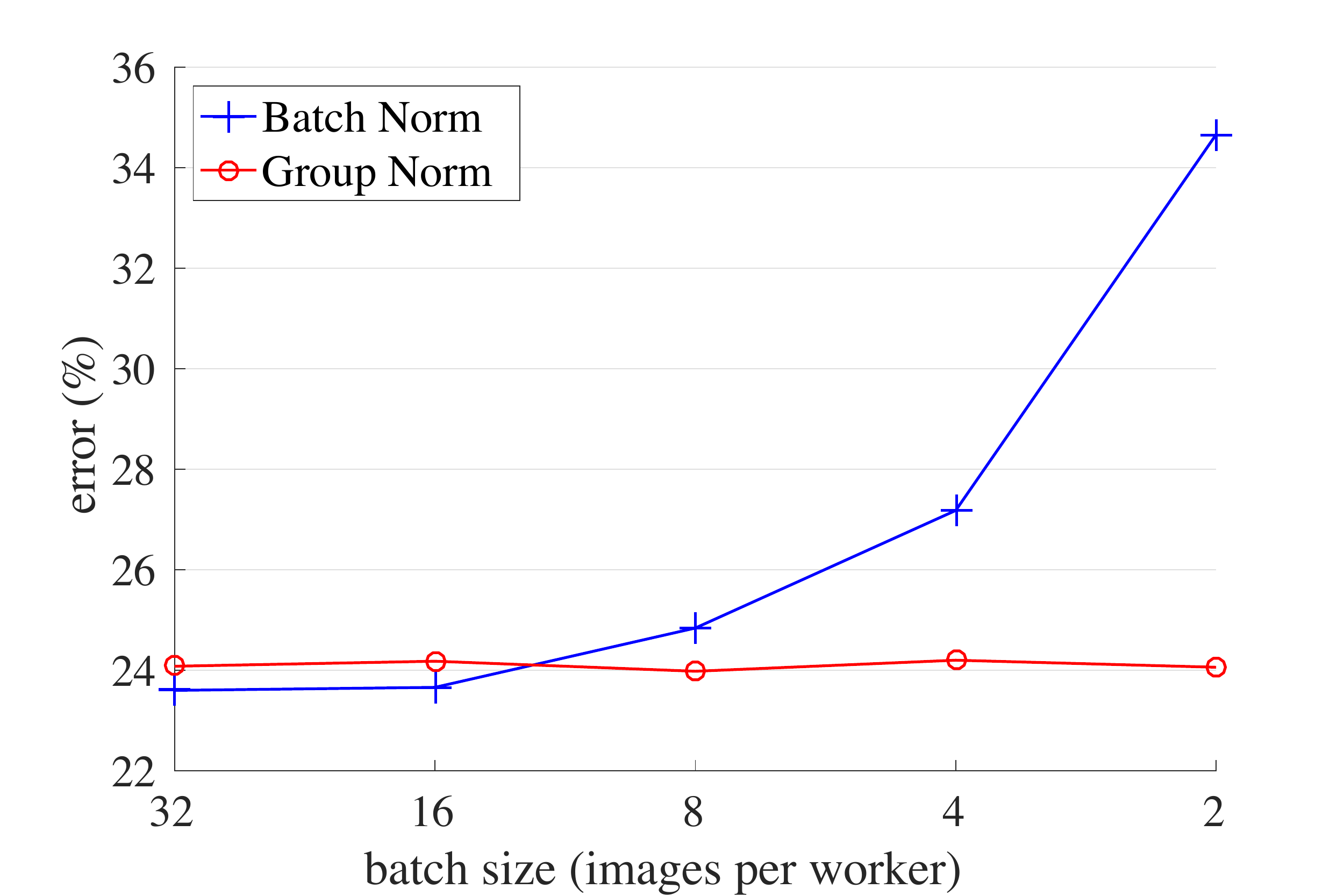}
\vspace{.1em}
\caption{\textbf{ImageNet classification error \vs batch sizes}. This is a ResNet-50 model trained in the ImageNet training set using 8 workers (GPUs), evaluated in the validation set.
}
\vspace{-1.5em}
\label{fig:teaser}
\end{figure}

Despite its great success, BN exhibits drawbacks that are also caused by its distinct behavior of normalizing along the batch dimension.
In particular, it is required for BN to work with a \emph{sufficiently large batch size} (\eg, 32 per worker\footnote{In the context of this paper, we use ``batch size'' to refer to the number of samples \emph{per worker} (\eg, GPU). BN's statistics are computed for each worker, but \emph{not} broadcast across workers, as is standard in many libraries.} \cite{Ioffe2015,Szegedy2016a,He2016}). %
A small batch leads to inaccurate estimation of the batch statistics, and \emph{reducing BN's batch size increases the model error dramatically} (Figure~\ref{fig:teaser}). As a result, many recent models \cite{Szegedy2016a,He2016,Szegedy2016,Huang2017,Xie2017} are trained with non-trivial batch sizes that are memory-consuming. The heavy reliance on BN's effectiveness to train models in turn prohibits people from exploring higher-capacity models that would be limited by memory.

The restriction on batch sizes is more demanding in computer vision tasks including detection \cite{Girshick2015,Ren2015,He2017}, segmentation \cite{Long2015,He2017}, video recognition \cite{Tran2015,Carreira2017}, and other high-level systems built on them. For example, the Fast/er and Mask R-CNN frameworks \cite{Girshick2015,Ren2015,He2017} use a batch size of 1 or 2 images because of higher resolution, where BN is ``frozen'' by transforming to a linear layer \cite{He2016}; in video classification with 3D convolutions \cite{Tran2015,Carreira2017}, the presence of spatial-temporal features introduces a trade-off between the temporal length and batch size. The usage of BN often requires these systems to compromise between the model design and batch sizes.

This paper presents Group Normalization (GN) as a simple alternative to BN. We notice that many classical features like SIFT \cite{Lowe2004} and HOG \cite{Dalal2005} are \emph{group-wise} features and involve \emph{group-wise normalization}. For example, a HOG vector is the outcome of several spatial cells where each cell is represented by a normalized orientation histogram.
Analogously, we propose GN as a layer that divides channels into groups and normalizes the features within each group (Figure~\ref{fig:all_norms}). 
GN does not exploit the batch dimension, and its computation is independent of batch sizes.

GN behaves very stably over a wide range of batch sizes (Figure~\ref{fig:teaser}).
With a batch size of 2 samples, GN has 10.6\% lower error than its BN counterpart for ResNet-50 \cite{He2016} in ImageNet \cite{Russakovsky2015}. With a regular batch size, GN is comparably good as BN (with a gap of $\app$0.5\%) and outperforms other normalization variants \cite{Ba2016,Ulyanov2016,Salimans2016}. Moreover, although the batch size may change, GN can naturally transfer from pre-training to fine-tuning.
GN shows improved results \vs its BN counterpart on Mask R-CNN for COCO object detection and segmentation \cite{Lin2014}, and on 3D convolutional networks for Kinetics video classification \cite{Kay2017}. The effectiveness of GN in ImageNet, COCO, and Kinetics demonstrates that GN is a competitive alternative to BN that has been dominant in these tasks.

There have been existing methods, such as Layer Normalization (LN) \cite{Ba2016} and Instance Normalization (IN) \cite{Ulyanov2016} (Figure~\ref{fig:all_norms}), that also avoid normalizing along the batch dimension.
These methods are effective for training sequential models (RNN/LSTM \cite{Rumelhart1986,Hochreiter1997}) or generative models (GANs \cite{Goodfellow2014,Isola2017}).
But as we will show by experiments, both LN and IN have limited success in visual recognition, for which GN presents better results.
Conversely, GN could be used in place of LN and IN and thus is applicable for sequential or generative models. This is beyond the focus of this paper, but it is suggestive for future research.

\section{Related Work}

\paragraph{Normalization.} It is well-known that normalizing the input data makes training faster \cite{LeCun1998}. To normalize hidden features, initialization methods \cite{LeCun1998,Glorot2010,He2015} have been derived based on strong assumptions of feature distributions, which can become invalid when training evolves.

Normalization layers in deep networks had been widely used before the development of BN. Local Response Normalization (LRN) \cite{Lyu2008,Jarrett2009,Krizhevsky2012} was a component in AlexNet \cite{Krizhevsky2012} and following models \cite{Zeiler2014,Sermanet2014,Szegedy2015}. Unlike recent methods \cite{Ioffe2015,Ba2016,Ulyanov2016}, LRN computes the statistics in a small neighborhood for each pixel.

Batch Normalization \cite{Ioffe2015} performs more global normalization along the batch dimension (and as importantly, it suggests to do this for all layers). 
But the concept of ``batch'' is not always present, or it may change from time to time. For example, batch-wise normalization is not legitimate at inference time, so the mean and variance are pre-computed from the training set \cite{Ioffe2015}, often by running average;
consequently, there is no normalization performed when testing. The pre-computed statistics may also change when the target data distribution changes \cite{Rebuffi2017}. These issues lead to inconsistency at training, transferring, and testing time. In addition, as aforementioned, reducing the batch size can have dramatic impact on the estimated batch statistics.

\begin{figure*}[t]
\centering
\includegraphics[width=.72\linewidth]{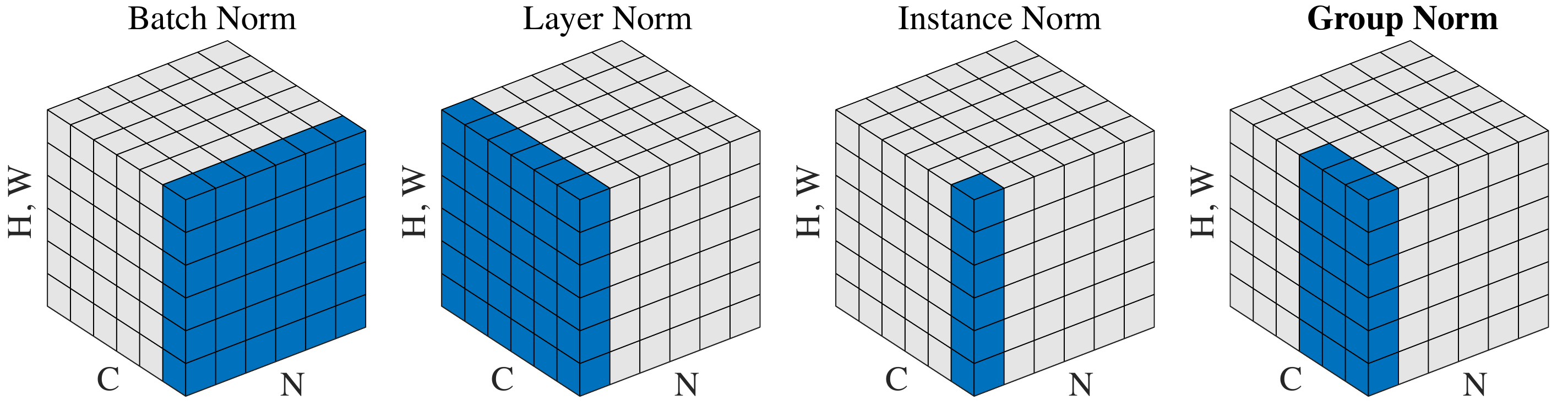}
\vspace{.5em}
\caption{\textbf{Normalization methods}. Each subplot shows a feature map tensor, with $N$ as the batch axis, $C$ as the channel axis, and $(H, W)$ as the spatial axes.
The pixels in blue are normalized by the same mean and variance, computed by aggregating the values of these pixels.
}
\label{fig:all_norms}
\vspace{-.5em}
\end{figure*}

Several normalization methods \cite{Ba2016,Ulyanov2016,Salimans2016,Arpit2016,Ren2017a} have been proposed to avoid exploiting the batch dimension. Layer Normalization (LN) \cite{Ba2016} operates along the channel dimension, and Instance Normalization (IN) \cite{Ulyanov2016} performs BN-like computation but only for each sample (Figure~\ref{fig:all_norms}).
Instead of operating on features, Weight Normalization (WN) \cite{Salimans2016} proposes to normalize the filter weights. These methods do not suffer from the issues caused by the batch dimension, but they have not been able to approach BN's accuracy in many visual recognition tasks.
We provide comparisons with these methods in context of the remaining sections.

\paragraph{Addressing small batches.} 
Ioffe \cite{Ioffe2017} proposes Batch Renormalization (BR) that alleviates BN's issue involving small batches. BR introduces two extra parameters that constrain the estimated mean and variance of BN within a certain range, reducing their drift when the batch size is small.
BR has better accuracy than BN in the small-batch regime. But BR is also batch-dependent, and when the batch size decreases its accuracy still degrades \cite{Ioffe2017}.

There are also attempts to \emph{avoid} using small batches. The object detector in \cite{Peng2018} performs synchronized BN whose mean and variance are computed across multiple GPUs.
However, this method does not solve the problem of small batches; instead, it migrates the algorithm problem to engineering and hardware demands, using a number of GPUs proportional to BN's requirements.
Moreover, the synchronized BN computation prevents using \emph{asynchronous} solvers (ASGD \cite{Dean2012}), a practical solution to large-scale training widely used in industry. These issues can limit the scope of using synchronized BN.

Instead of addressing the batch statistics computation (\eg, \cite{Ioffe2017,Peng2018}), our normalization method inherently avoids this computation.

%%%
%%%

\paragraph{Group-wise computation.}
\emph{Group convolutions} have been presented by AlexNet \cite{Krizhevsky2012} for distributing a model into two GPUs. The concept of \emph{groups} as a dimension for model design has been more widely studied recently. The work of ResNeXt  \cite{Xie2017} investigates the trade-off between depth, width, and groups, and it suggests that a larger number of groups can improve accuracy under similar computational cost. MobileNet \cite{Howard2017} and Xception \cite{Chollet2017} exploit \emph{channel-wise} (also called ``depth-wise'') convolutions, which are group convolutions with a group number equal to the channel number. ShuffleNet \cite{Zhang2018} proposes a channel shuffle operation that permutes the axes of grouped features. These methods all involve dividing the channel dimension into groups.
Despite the relation to these methods, GN does \emph{not} require group convolutions. GN is a generic layer, as we evaluate in standard ResNets \cite{He2016}.

\section{Group Normalization}

The channels of visual representations are not entirely independent.
Classical features of SIFT \cite{Lowe2004}, HOG \cite{Dalal2005}, and GIST \cite{Oliva2001} are \emph{group-wise} representations by design, where each group of channels is constructed by some kind of histogram.
These features are often processed by \emph{group-wise normalization} over each histogram or each orientation. Higher-level features such as VLAD \cite{Jegou2010} and Fisher Vectors (FV) \cite{Perronnin2007} are also group-wise features where a group can be thought of as the sub-vector computed with respect to a cluster.

Analogously, it is not necessary to think of deep neural network features as unstructured vectors. For example, for conv$_1$ (the first convolutional layer) of a network, it is reasonable to expect a filter and its horizontal flipping to exhibit similar distributions of filter responses on natural images. If conv$_1$ happens to approximately learn this pair of filters, or if the horizontal flipping (or other transformations) is made into the architectures by design \cite{Dieleman2016,Cohen2016}, then the corresponding channels of these filters can be normalized together.

The higher-level layers are more abstract and their behaviors are not as intuitive. However, in addition to orientations (SIFT \cite{Lowe2004}, HOG \cite{Dalal2005}, or \cite{Dieleman2016,Cohen2016}), there are many factors that could lead to grouping, \eg, frequency, shapes, illumination, textures. Their coefficients can be interdependent.
In fact, a well-accepted computational model in neuroscience is to normalize across the cell responses \cite{Heeger1992,Schwartz2001,Simoncelli2001,Carandini2012}, ``with various receptive-field centers (covering the visual field) and with various spatiotemporal frequency tunings'' (p183, \cite{Heeger1992}); this can happen not only  in the primary visual cortex, but also ``throughout the visual system'' \cite{Carandini2012}.
Motivated by these works, we propose new generic group-wise normalization for deep neural networks.

\subsection{Formulation}

We first describe a general formulation of feature normalization, and then present GN in this formulation.
A family of feature normalization methods, including BN, LN, IN, and GN,
perform the following computation:
\begin{equation}\label{eq:norm}
\hat{x}_i = \frac{1}{\sigma_i}(x_i - \mu_i).
\end{equation}
Here $x$ is the feature computed by a layer, and $i$ is an index. In the case of 2D images, $i=(i_N, i_C, i_H, i_W)$ is a 4D vector indexing the features in $(N, C, H, W)$ order, where $N$ is the batch axis, $C$ is the channel axis, and $H$ and $W$ are the spatial height and width axes.

$\mu$ and $\sigma$ in (\ref{eq:norm}) are the mean and standard deviation (std) computed by:
\begin{equation}\label{eq:mu_sigma}
\mu_i = \frac{1}{m}\sum_{k\in \mathcal{S}_i} x_k, \quad \sigma_i = \sqrt{ \frac{1}{m}\sum_{k\in \mathcal{S}_i} (x_k-\mu_i)^2 + \epsilon},
\end{equation}
with $\epsilon$ as a small constant. $\mathcal{S}_i$ is the set of pixels in which the mean and std are computed, and $m$ is the size of this set. Many types of feature normalization methods mainly differ in how the set $\mathcal{S}_i$ is defined (Figure~\ref{fig:all_norms}), discussed as follows.

In \textbf{Batch Norm} \cite{Ioffe2015}, the set $\mathcal{S}_i$ is defined as:
\begin{equation}\label{eq:set_bn}
\mathcal{S}_i = \{k~|~k_C=i_C\},
\end{equation}
where $i_C$ (and $k_C$) denotes the sub-index of $i$ (and $k$) along the $C$ axis. This means that the pixels sharing the same channel index are normalized together, \ie, for each channel, BN computes $\mu$ and $\sigma$ along the $(N, H, W)$ axes.
In \textbf{Layer Norm} \cite{Ba2016}, the set is:
\begin{equation}\label{eq:set_ln}
\mathcal{S}_i = \{k~|~k_N=i_N\},
\end{equation}
meaning that LN computes $\mu$ and $\sigma$ along the $(C, H, W)$ axes for each sample.
In \textbf{Instance Norm} \cite{Ulyanov2016}, the set is:
\begin{equation}\label{eq:set_in}
\mathcal{S}_i = \{k~|~k_N=i_N, k_C=i_C\}.
\end{equation}
meaning that IN computes $\mu$ and $\sigma$ along the $(H, W)$ axes for each sample and each channel.
The relations among BN, LN, and IN are in Figure~\ref{fig:all_norms}.

As in \cite{Ioffe2015}, all methods of BN, LN, and IN learn a per-channel linear transform to compensate for the possible lost of representational ability:
\begin{equation}\label{eq:gamma_beta}
y_i = \gamma \hat{x}_i + \beta,
\end{equation}
where $\gamma$ and $\beta$ are trainable scale and shift (indexed by $i_C$ in all case, which we omit for simplifying notations).

\paragraph{Group Norm.}
Formally, a Group Norm layer computes $\mu$ and $\sigma$ in a set $\mathcal{S}_i$ defined as:
\begin{equation}\label{eq:set_gn}
\mathcal{S}_i = \{k~|~k_N=i_N, \lfloor \frac{k_C}{C/G} \rfloor=\lfloor \frac{i_C}{C/G} \rfloor\}.
\end{equation}
Here $G$ is the number of groups, which is a pre-defined hyper-parameter ($G=32$ by default). $C/G$ is the number of channels per group.
$\lfloor\cdot\rfloor$ is the floor operation, and ``$\lfloor \frac{k_C}{C/G} \rfloor=\lfloor \frac{i_C}{C/G}\rfloor$'' means that the indexes $i$ and $k$ are in the same group of channels, assuming each group of channels are stored in a sequential order along the $C$ axis.
GN computes $\mu$ and $\sigma$ along the $(H, W)$ axes and along a group of $\frac{C}{G}$ channels.
The computation of GN is illustrated in Figure~\ref{fig:all_norms} (rightmost), which is a simple case of 2 groups ($G=2$) each having 3 channels.

Given $\mathcal{S}_i$ in Eqn.(\ref{eq:set_gn}), a GN layer is defined by Eqn.(\ref{eq:norm}), (\ref{eq:mu_sigma}), and (\ref{eq:gamma_beta}). Specifically, the pixels in the same group are normalized together by the same $\mu$ and $\sigma$. GN also learns the per-channel $\gamma$ and $\beta$.

\paragraph{Relation to Prior Work.} LN, IN, and GN all perform independent computations along the batch axis.
The two extreme cases of GN are equivalent to LN and IN (Figure~\ref{fig:all_norms}).

\begin{figure*}[t]
\centering
\includegraphics[width=.36\linewidth]{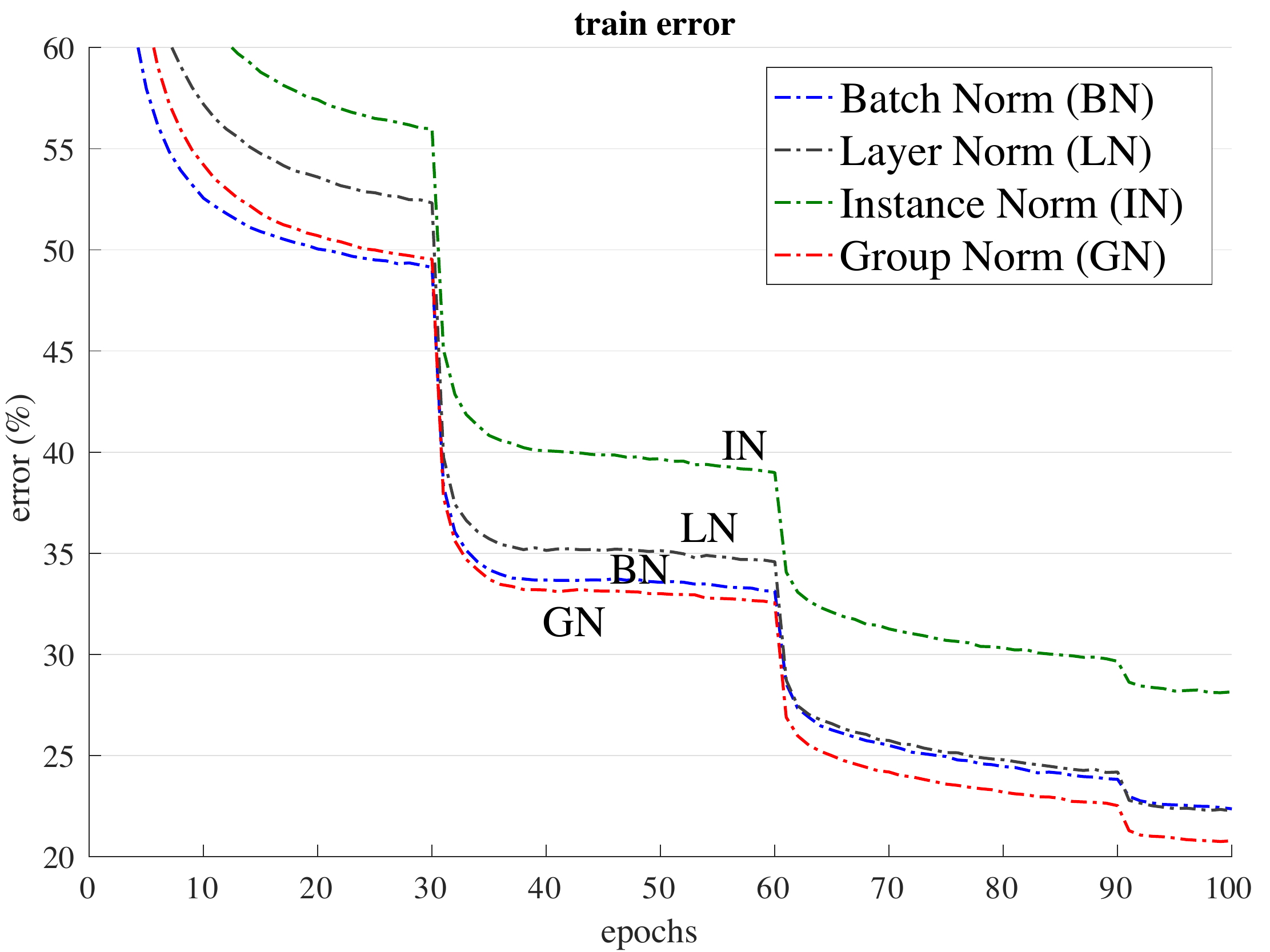}
\includegraphics[width=.36\linewidth]{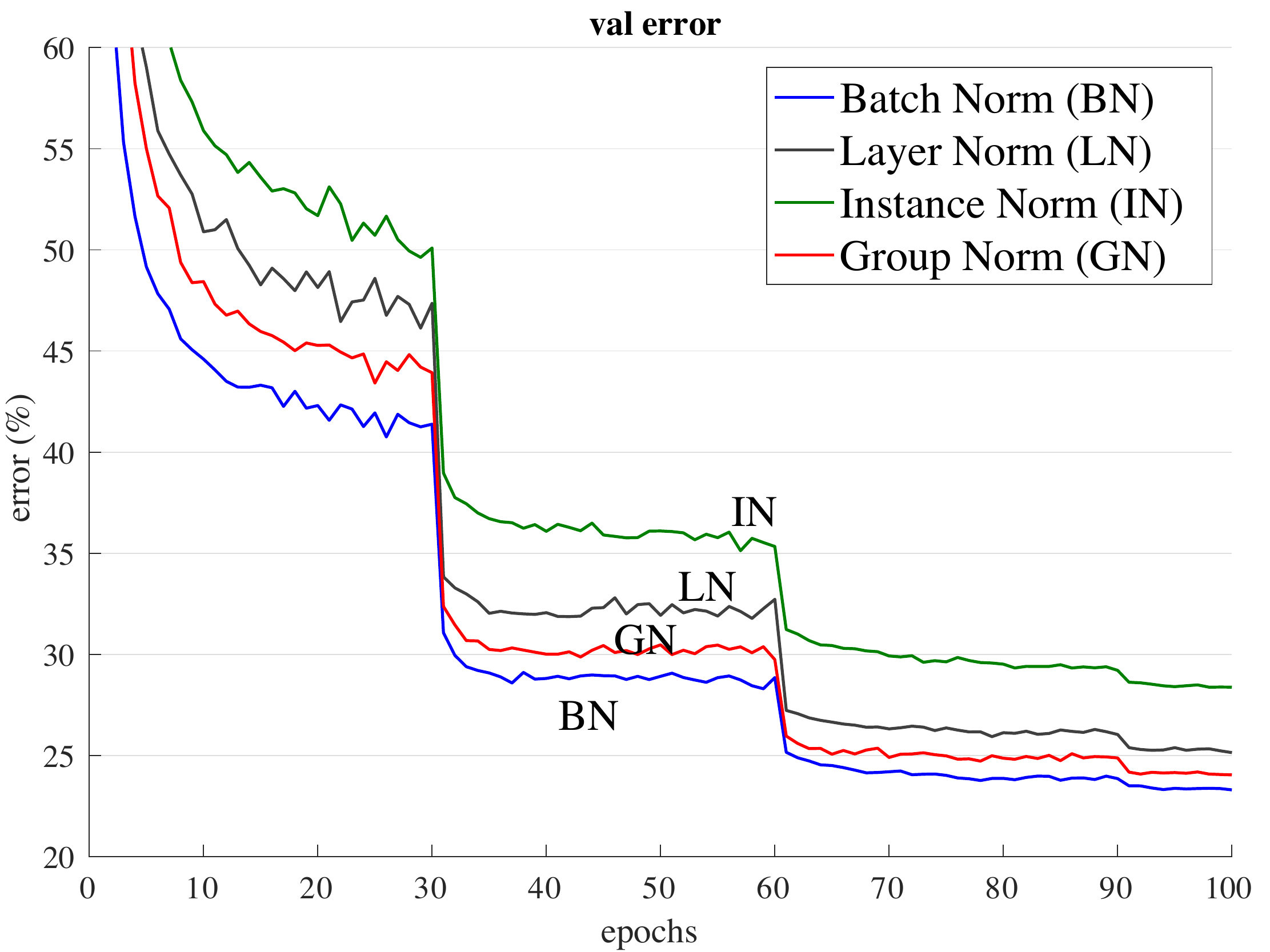}
\vspace{.3em}
\caption{\textbf{Comparison of error curves} with a batch size of \textbf{32 images}/GPU. We show the ImageNet training error (left) and validation error (right) \vs numbers of training epochs. The model is ResNet-50.}
\label{fig:curves_all_norms}
\vspace{-.5em}
\end{figure*}

\vspace{.5em}\noindent
\emph{Relation to Layer Normalization} \cite{Ba2016}. GN becomes LN if we set the group number as $G=1$. LN assumes \emph{all} channels in a layer make ``similar contributions'' \cite{Ba2016}. Unlike the case of fully-connected layers studied in \cite{Ba2016}, this assumption can be less valid with the presence of convolutions, as discussed in \cite{Ba2016}. GN is less restricted than LN, because each group of channels (instead of all of them) are assumed to subject to the shared mean and variance; the model still has flexibility of learning a different distribution for each group. This leads to improved representational power of GN over LN, as shown by the lower training and validation error in experiments (Figure~\ref{fig:curves_all_norms}).

\vspace{.5em}\noindent
\emph{Relation to Instance Normalization} \cite{Ulyanov2016}. GN becomes IN if we set the group number as $G=C$ (\ie, one channel per group). But IN can only rely on the spatial dimension for computing the mean and variance and it misses the opportunity of exploiting the channel dependence.

\lstset{
  backgroundcolor=\color{white},
  basicstyle=\fontsize{7.5pt}{8.5pt}\fontfamily{lmtt}\selectfont,
  columns=fullflexible,
  breaklines=true,
  captionpos=b,
  commentstyle=\fontsize{8pt}{9pt}\color{codegray},
  keywordstyle=\fontsize{8pt}{9pt}\color{codegreen},
  stringstyle=\fontsize{8pt}{9pt}\color{codeblue},
  frame=tb,
  otherkeywords = {self},
}
\begin{figure}[t]
\tiny
%    # as reference: the following 2 lines implement BN
%    # mean, var = tf.nn.moments(x, [0, 2, 3], keep_dims=True)
%    # x = (x - mean) / tf.sqrt(var + epsilon)
\begin{lstlisting}[language=python]
def GroupNorm(x, gamma, beta, G, eps=1e-5):
    # x: input features with shape [N,C,H,W]
    # gamma, beta: scale and offset, with shape [1,C,1,1]
    # G: number of groups for GN

    N, C, H, W = x.shape
    x = tf.reshape(x, [N, G, C // G, H, W])

    mean, var = tf.nn.moments(x, [2, 3, 4], keep_dims=True)
    x = (x - mean) / tf.sqrt(var + eps)

    x = tf.reshape(x, [N, C, H, W])

    return x * gamma + beta
\end{lstlisting}
\caption{Python code of Group Norm based on TensorFlow.}
\label{fig:code}
\vspace{-1em}
\end{figure}

%% pytorch
%\lstset{
%  backgroundcolor=\color{white},
%  basicstyle=\fontsize{8pt}{10pt}\fontfamily{lmtt}\selectfont,
%  columns=fullflexible,
%  breaklines=true,
%  captionpos=b,
%  commentstyle=\fontsize{8pt}{9pt}\color{codegray},
%  keywordstyle=\fontsize{8pt}{9pt}\color{codegreen},
%  stringstyle=\fontsize{8pt}{9pt}\color{codeblue},
%  frame=tb,
%  otherkeywords = {self},
%}
%\begin{figure}[t]
%\tiny
%%    # as reference: the following 2 lines implement BN
%%    # mean, var = tf.nn.moments(x, [0, 2, 3], keep_dims=True)
%%    # x = (x - mean) / tf.sqrt(var + epsilon)
%\begin{lstlisting}[language=python]
%# GroupNorm
%def groupnorm(x, G, gamma, beta, eps=1e-5):
%    # x: input features with shape [N,C,H,W]
%    # gamma, beta: learnable scale and offset, with shape [1,C,1,1]
%    # G: number of groups for GN
%
%    N, C, H, W = x.shape
%    x = x.view(N * G, -1)  # reshape to [N*G, C//G*H*W]
%    mean = x.mean(1, keepdim=True)
%    var = x.var(1, keepdim=True)
%    x = (x - mean) / (var + eps).sqrt()
%    return gamma * x.view([N, C, H, W]) + beta
%\end{lstlisting}
%\vspace{-1em}
%\caption{Python code of Group Norm based on PyTorch. Here the function {\fontfamily{lmtt}\selectfont x.mean} and {\fontfamily{lmtt}\selectfont x.std} computes the mean and std by aggregating along the specified axes.}
%\label{fig:code}
%\vspace{-1em}
%\end{figure}

\subsection{Implementation}

GN can be easily implemented by a few lines of code in PyTorch \cite{Paszke2017} and TensorFlow \cite{Abadi2016} where automatic differentiation is supported. Figure~\ref{fig:code} shows the code based on TensorFlow. In fact, we only need to specify how the mean and variance (``moments'') are computed, along the appropriate axes as defined by the normalization method.

%%% -----------
%%%

%%% -----------
\begin{figure*}[t]
\centering
\includegraphics[width=.36\linewidth]{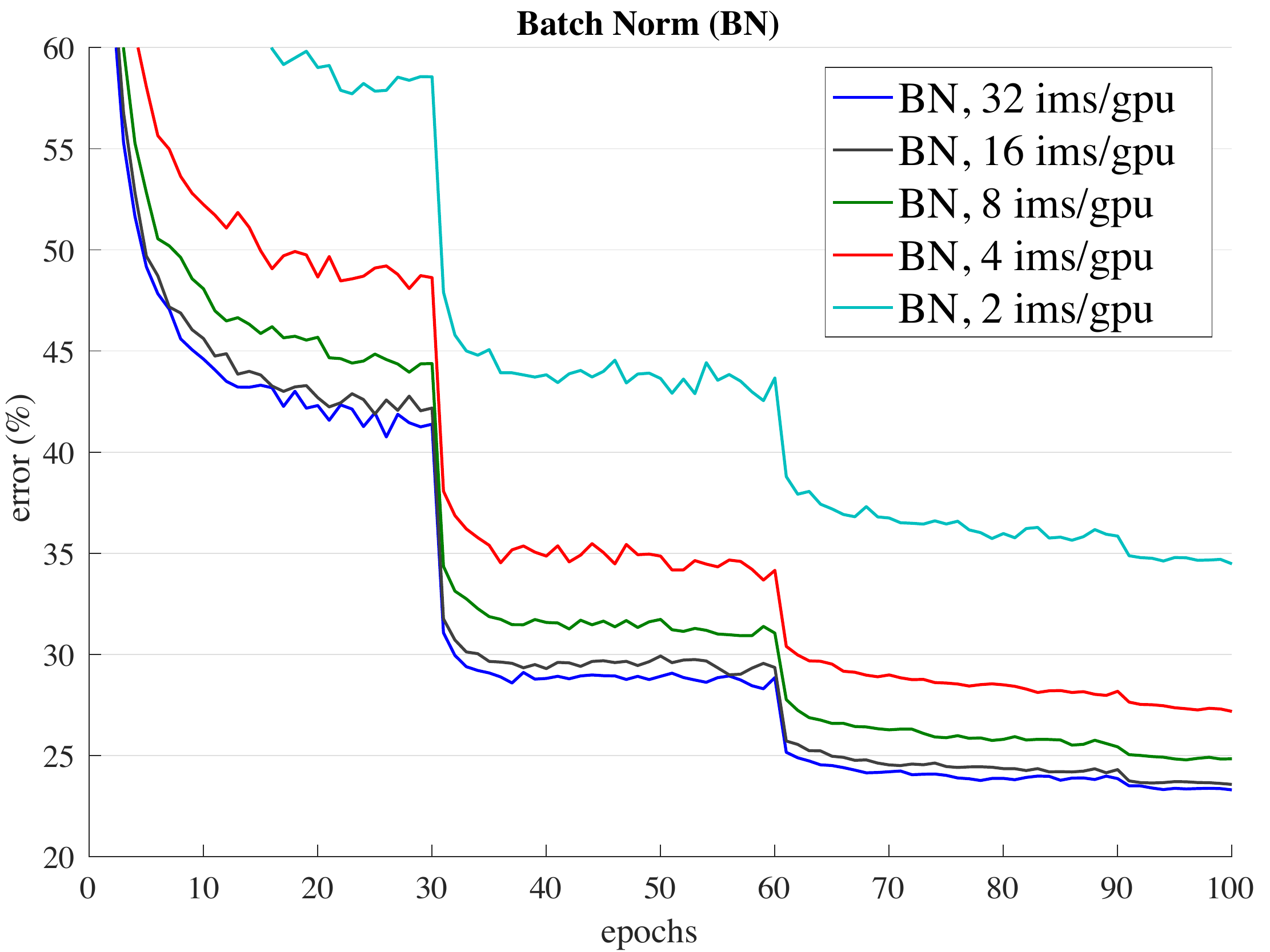}
\includegraphics[width=.36\linewidth]{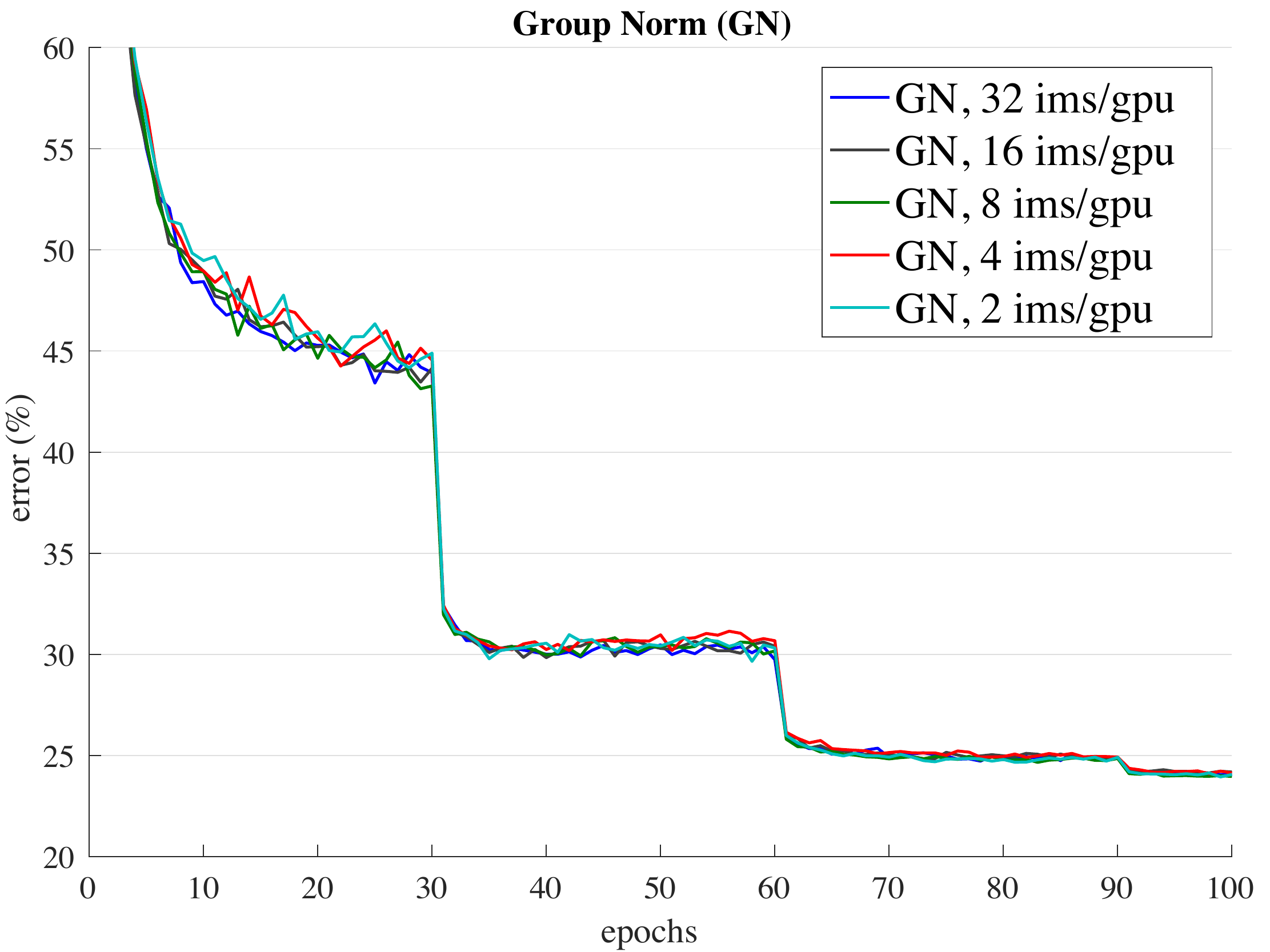}
\vspace{.3em}
\caption{\textbf{Sensitivity to batch sizes}: ResNet-50's validation error of BN (left) and GN (right) trained with 32, 16, 8, 4, and 2 images/GPU.}
\label{fig:curves_mbsize}
\end{figure*}
%%%

\section{Experiments}

\subsection{Image Classification in ImageNet}

We experiment in the ImageNet classification dataset \cite{Russakovsky2015} with 1000 classes. We train on the $\app$1.28M training images and evaluate on the 50,000 validation images, using the ResNet models \cite{He2016}.

\paragraph{Implementation details.} As standard practice \cite{He2016,Gross2016}, we use 8 GPUs to train all models, and the batch mean and variance of BN are computed \emph{within} each GPU. We use the method of \cite{He2015} to initialize all convolutions for all models. We use 1 to initialize all $\gamma$ parameters, except for each residual block's last normalization layer where we initialize $\gamma$ by 0 following \cite{Goyal2017} (such that the initial state of a residual block is identity). We use a weight decay of 0.0001 for all weight layers, including $\gamma$ and $\beta$ (following \cite{Gross2016} but unlike \cite{He2016,Goyal2017}).
We train 100 epochs for all models, and decrease the learning rate by 10$\times$ at 30, 60, and 90 epochs. During training, we adopt the data augmentation of \cite{Szegedy2015} as implemented by \cite{Gross2016}.
We evaluate the top-1 classification error on the center crops of 224$\times$224 pixels in the validation set. To reduce random variations, we report the median error rate of the final 5 epochs \cite{Goyal2017}. Other implementation details follow \cite{Gross2016}.

Our baseline is the ResNet trained with BN \cite{He2016}. To compare with LN, IN, and GN, we replace BN with the specific variant. We use the same hyper-parameters for all models. We set $G=32$ for GN by default.

\paragraph{Comparison of feature normalization methods.}
We first experiment with a regular batch size of \textbf{32 images} (per GPU) \cite{Ioffe2015,He2016}. BN works successfully in this regime, so this is a strong baseline to compare with.
Figure~\ref{fig:curves_all_norms} shows the error curves, and Table~\ref{tab:all_norms} shows the final results.

%%% -----------
\renewcommand\arraystretch{1.05}
\setlength{\tabcolsep}{8pt}
\begin{table}[t]
\centering
\small
\begin{tabular}{c|cccc}
\hline
 & BN & LN & IN & GN \\
\shline
val error & \textbf{23.6} & 25.3 & 28.4 & 24.1 \\
$\triangle$ (\vs BN) & - & \emph{1.7} & \emph{4.8} & \textbf{\emph{0.5}} \\
\hline
\end{tabular}
\vspace{1em}
\caption{\textbf{Comparison of error rates} (\%) of ResNet-50 in the ImageNet validation set, trained
with a batch size of \textbf{32 images}/GPU. The error curves are in Figure~\ref{fig:curves_all_norms}.}\label{tab:all_norms}
\vspace{-1.5em}
\end{table}
%%% -----------

Figure~\ref{fig:curves_all_norms} shows that \emph{all} of these normalization methods are able to converge. LN has a small degradation of 1.7\% comparing with BN. This is an encouraging result, as it suggests that normalizing along \emph{all} channels (as done by LN) of a \emph{convolutional} network is reasonably good. IN also makes the model converge, but is 4.8\% worse than BN.\footnote{For completeness, we have also trained ResNet-50 with WN \cite{Salimans2016}, which is \emph{filter} (instead of \emph{feature}) normalization. WN's result is 28.2\%. }

In this regime where BN works well, GN is able to approach BN's accuracy, with a decent degradation of 0.5\% in the validation set. Actually, Figure~\ref{fig:curves_all_norms} (left) shows that GN has \emph{lower training error} than BN, indicating that GN is effective for easing \emph{optimization}. The slightly higher validation error of GN implies that GN loses some regularization ability of BN. This is understandable, because BN's mean and variance computation introduces uncertainty caused by the stochastic batch sampling, which helps regularization \cite{Ioffe2015}. This uncertainty is missing in GN (and LN/IN). But it is possible that GN combined with a suitable regularizer will improve results. This can be a future research topic.

%%%
\renewcommand\arraystretch{1.05}
\setlength{\tabcolsep}{8pt}
\begin{table}[t]
\centering
\small
\begin{tabular}{c|ccccc}
\hline
batch size & 32 & 16 & 8 & 4 & 2\\
\shline
BN & \textbf{23.6} & \textbf{23.7} & 24.8 & 27.3 & 34.7 \\
GN & 24.1 & 24.2 & \textbf{24.0} & \textbf{24.2} & \textbf{24.1} \\
$\triangle$ & \emph{0.5} & \emph{0.5} & \emph{-0.8} & \emph{-3.1} & \emph{-10.6} \\
\hline
\end{tabular}
\vspace{1em}
\caption{\textbf{Sensitivity to batch sizes}. We show ResNet-50's validation error (\%) in ImageNet. The last row shows the differences between BN and GN.
The error curves are in Figure~\ref{fig:curves_mbsize}. This table is visualized in Figure~\ref{fig:teaser}.}\label{tab:mbsize}
\vspace{-1em}
\end{table}
%%% -----------

%%% -----------
\begin{figure*}[t]
\begin{minipage}[c]{0.84\linewidth}
\centering
\includegraphics[width=.31\linewidth]{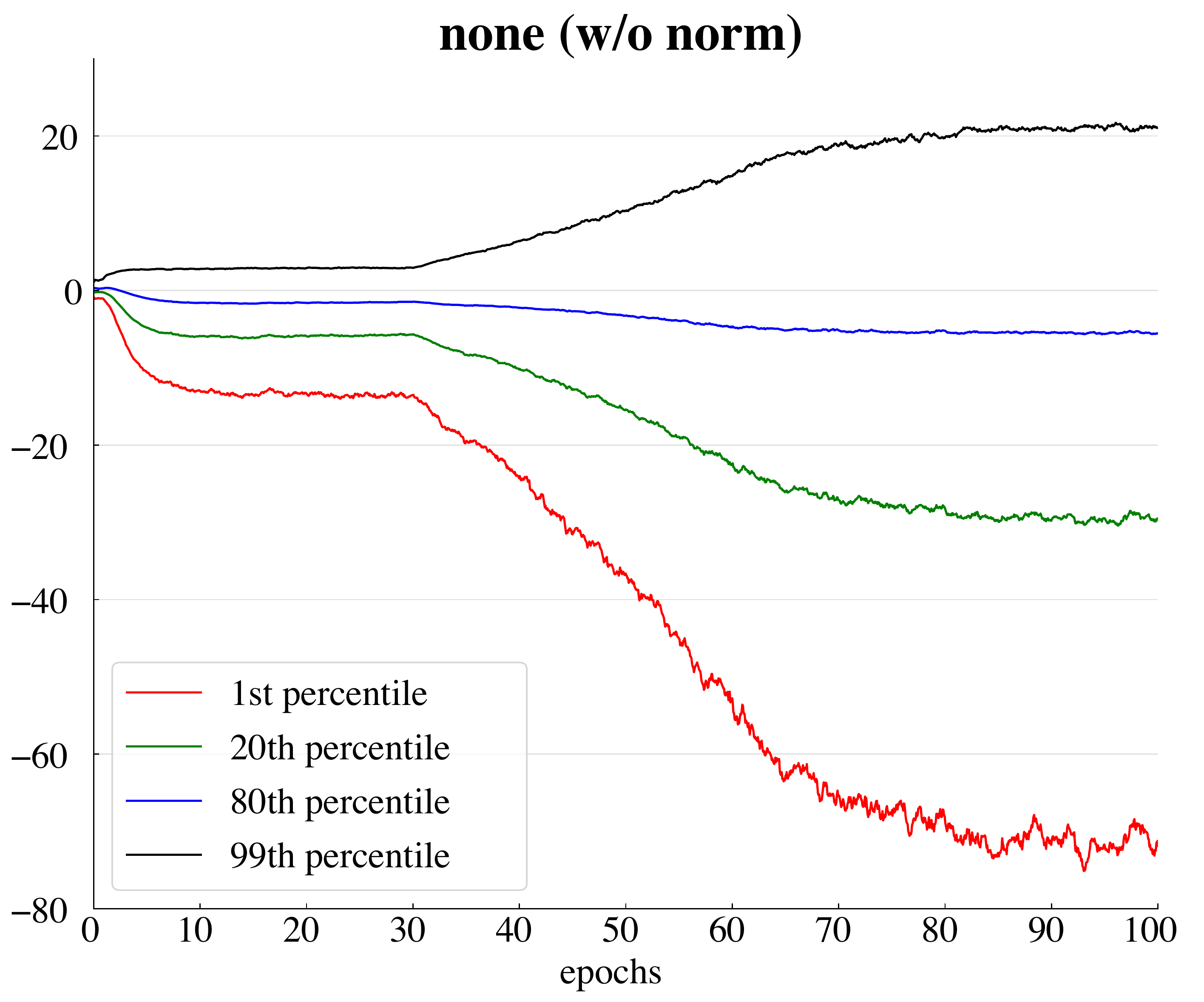}
\includegraphics[width=.31\linewidth]{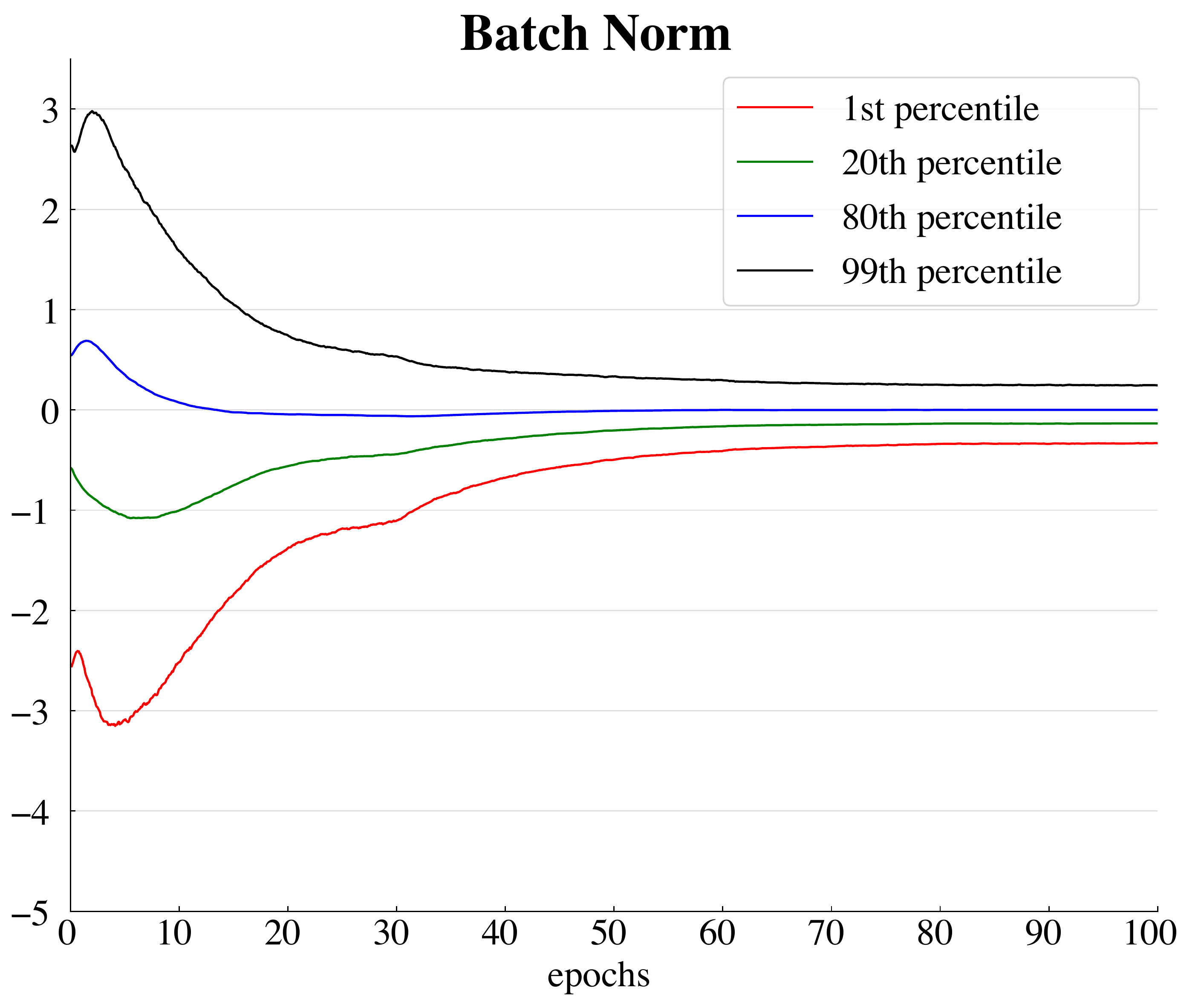}
\includegraphics[width=.31\linewidth]{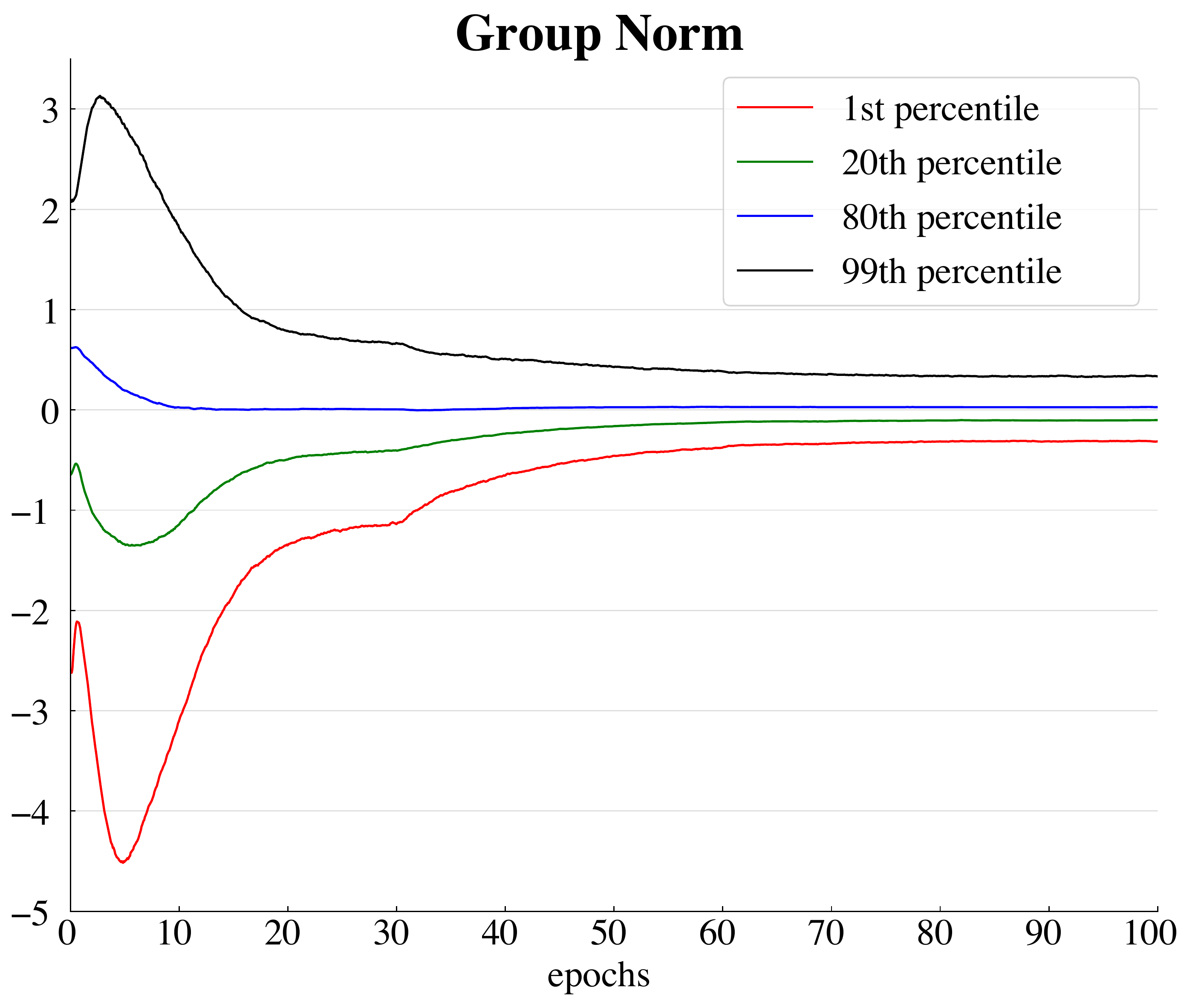}
\end{minipage}
\begin{minipage}[c]{0.15\linewidth}
\renewcommand\arraystretch{1.05}
\setlength{\tabcolsep}{4pt}
\small
\flushleft
\begin{tabular}{c|c}
\hline
  & error \\
\shline
none & 29.2 \\ 
BN & 28.0 \\
GN & \textbf{27.6} \\
\hline
\end{tabular}
\end{minipage}
\vspace{.5em}
\caption{\textbf{Evolution of feature distributions} of conv$_{5\_3}$'s output (before normalization and ReLU) from VGG-16, shown as the \{1, 20, 80, 99\} percentile of responses. The table on the right shows the ImageNet validation error (\%). Models are trained with 32 images/GPU.}
\label{fig:vgg}
\vspace{-1em}
\end{figure*}
%%% -----------

\paragraph{Small batch sizes.}
Although BN benefits from the stochasticity under some situations, its error increases when the batch size becomes smaller and the uncertainty gets bigger. We show this in Figure~\ref{fig:teaser}, Figure~\ref{fig:curves_mbsize}, and Table~\ref{tab:mbsize}.

We evaluate batch sizes of 32, 16, 8, 4, 2 images per GPU. In all cases, the BN mean and variance are computed within each GPU and not synchronized. All models are trained in 8 GPUs. In this set of experiments, we adopt the linear learning rate scaling rule \cite{Krizhevsky2014,Bottou2016,Goyal2017} to adapt to batch size changes --- we use a learning rate of 0.1 \cite{He2016} for the batch size of 32, and 0.1$N/$32 for a batch size of $N$. This linear scaling rule works well for BN if the total batch size changes (by changing the number of GPUs) but the per-GPU batch size does not change \cite{Goyal2017}.
We keep the same number of training epochs for all cases (Figure~\ref{fig:curves_mbsize}, x-axis). All other hyper-parameters are unchanged.

Figure~\ref{fig:curves_mbsize} (left) shows that BN's error becomes considerably higher with small batch sizes. GN's behavior is more stable and insensitive to the batch size. Actually, Figure~\ref{fig:curves_mbsize} (right) shows that GN has very similar curves (subject to random variations) across a wide range of batch sizes from 32 to 2. In the case of a batch size of 2, GN has \textbf{10.6\%} lower error rate than its BN counterpart (24.1\% \vs 34.7\%).

These results indicate that the batch mean and variance estimation can be overly stochastic and inaccurate, especially when they are computed over 4 or 2 images. However, this stochasticity disappears if the statistics are computed from 1 image, in which case BN becomes similar to IN at training time. We see that IN has a better result (28.4\%) than BN with a batch size of 2 (34.7\%).

The robust results of GN in Table~\ref{tab:mbsize} demonstrate GN's strength. It allows to remove the batch size constraint imposed by BN, which can give considerably more memory (\eg, 16$\times$ or more).
This will make it possible to train higher-capacity models that would be otherwise bottlenecked by memory limitation. We hope this will create new opportunities in architecture design.

%%% -----------
\renewcommand\arraystretch{1.05}
\setlength{\tabcolsep}{2pt}
\begin{table}[t]
\small
\centering
\begin{tabular}{x{22}x{22}x{22}x{22}x{22}x{22}|x{32}}
 \multicolumn{7}{c}{\# groups ($G$)} \\ \hline
 64 & 32 & 16 & 8 & 4 & 2 & 1 (=LN) \\
\shline
 24.6 & \textbf{24.1} & 24.6 & 24.4 & 24.6 & 24.7 & 25.3 \\
 \emph{0.5} & - & \emph{0.5} & \emph{0.3} & \emph{0.5} & \emph{0.6} & \emph{1.2} \\
\hline
\end{tabular}
\\~\vspace{1em}\\
%%%
\centering
\begin{tabular}{x{22}x{22}x{22}x{22}x{22}x{22}|x{32}}
 \multicolumn{7}{c}{\# channels per group}  \\ \hline
64 &  32 & 16 & 8 & 4 & 2 & 1 (=IN) \\
\shline
24.4 & 24.5 & \textbf{24.2} & 24.3 & 24.8 & 25.6 & 28.4 \\
\emph{0.2} & \emph{0.3} & - & \emph{0.1} & \emph{0.6} & \emph{1.4} & \emph{4.2} \\
\hline
\end{tabular}
%\end{minipage}
\vspace{1em}
\caption{\textbf{Group division}. We show ResNet-50's validation error (\%) in ImageNet, trained with 32 images/GPU. (Top): a given number of groups. (Bottom): a given number of channels per group. The last rows show the differences with the best number.}\label{tab:groups}
\vspace{-1em}
\end{table}
%%% -----------

\paragraph{Comparison with Batch Renorm (BR).}
BR \cite{Ioffe2017} introduces two extra parameters ($r$ and $d$ in \cite{Ioffe2017}) that constrain the estimated mean and variance of BN. Their values are controlled by $r_\text{max}$ and $d_\text{max}$. To apply BR to ResNet-50, we have carefully chosen these hyper-parameters, and found that $r_\text{max}=1.5$ and $d_\text{max}=0.5$ work best for ResNet-50. With a batch size of 4, ResNet-50 trained with BR has an error rate of 26.3\%. This is better than BN's 27.3\%, but still 2.1\% higher than GN's 24.2\%.

\paragraph{Group division.} Thus far all presented GN models are trained with a group number of $G=32$. Next we evaluate different ways of dividing into groups.
With a given fixed group number, GN performs reasonably well for all values of $G$ we studied (Table~\ref{tab:groups}, top panel). In the extreme case of $G=1$, GN is equivalent to LN, and its error rate is higher than all cases of $G>1$ studied.

We also evaluate fixing the number of channels per group (Table~\ref{tab:groups}, bottom panel). Note that because the layers can have different channel numbers, the group number $G$ can change across layers in this setting. In the extreme case of 1 channel per group, GN is equivalent to IN. Even if using as few as 2 channels per group, GN has substantially lower error than IN (25.6\% \vs 28.4\%). This result shows the effect of grouping channels when performing normalization.

\paragraph{Deeper models.} We have also compared GN with BN on ResNet-101 \cite{He2016}. With a batch size of 32, our BN baseline of ResNet-101 has 22.0\% validation error, and the GN counterpart has 22.4\%, slightly worse by 0.4\%. With a batch size of 2, GN ResNet-101's error is 23.0\%. This is still a decently stable result considering the very small batch size, and it is 8.9\% better than the BN counterpart's 31.9\%.

\paragraph{Results and analysis of VGG models.} To study GN/BN compared to \emph{no normalization}, we consider VGG-16 \cite{Simonyan2015} that can be healthily trained without normalization layers. We apply BN or GN right after each convolutional layer.
Figure~\ref{fig:vgg} shows the evolution of the feature distributions of conv$_{5\_3}$ (the last convolutional layer). GN and BN behave \emph{qualitatively similar}, while being substantially different with the variant that uses no normalization; this phenomenon is also observed for all other convolutional layers. This comparison suggests that performing normalization is essential for controlling the distribution of features.

For VGG-16, GN is \emph{better} than BN by 0.4\% (Figure~\ref{fig:vgg}, right). This possibly implies that VGG-16 benefits less from BN's regularization effect, and GN (that leads to lower training error) is superior to BN in this case.

\subsection{Object Detection and Segmentation in COCO}

Next we evaluate fine-tuning the models for transferring to object detection and segmentation. These computer vision tasks in general benefit from higher-resolution input, so the batch size tends to be small in common practice (1 or 2 images/GPU \cite{Girshick2015,Ren2015,He2017,Lin2017a}). As a result, BN is turned into a \emph{linear} layer $y=\frac{\gamma}{\sigma}(x-\mu)+\beta$ where $\mu$ and $\sigma$ are pre-computed from the pre-trained model and frozen \cite{He2016}. We denote this as BN\f, which in fact performs no normalization during fine-tuning. We have also tried a variant that fine-tunes BN (normalization is performed and not frozen) and found it works poorly (reducing $\app$6 AP with a batch size of 2), so we ignore this variant.

We experiment on the Mask R-CNN baselines \cite{He2017}, implemented in the publicly available codebase of \emph{Detectron} \cite{Detectron2018}. We use the end-to-end variant with the same hyper-parameters as in \cite{Detectron2018}. 
We replace BN\f with GN during fine-tuning, using the corresponding models pre-trained from ImageNet.\footnote{Detectron \cite{Detectron2018} uses pre-trained models provided by the authors of \cite{He2016}. For fair comparisons, we instead use the models pre-trained in this paper. The object detection and segmentation accuracy is statistically similar between these pre-trained models.}
During fine-tuning, we use a weight decay of 0 for the $\gamma$ and $\beta$ parameters, which is important for good detection results when $\gamma$ and $\beta$ are being tuned. We fine-tune with a batch size of 1 image/GPU and 8 GPUs.

The models are trained in the COCO \texttt{train2017} set and evaluated in the COCO \texttt{val2017} set (a.k.a \texttt{minival}). We report the standard COCO metrics of Average Precision (AP), AP$_\text{50}$, and AP$_\text{75}$, for bounding box detection (AP$^\text{bbox}$) and instance segmentation (AP$^\text{mask}$).

%%% -----------
\begin{figure*}[t]
\centering
\includegraphics[width=.4\linewidth]{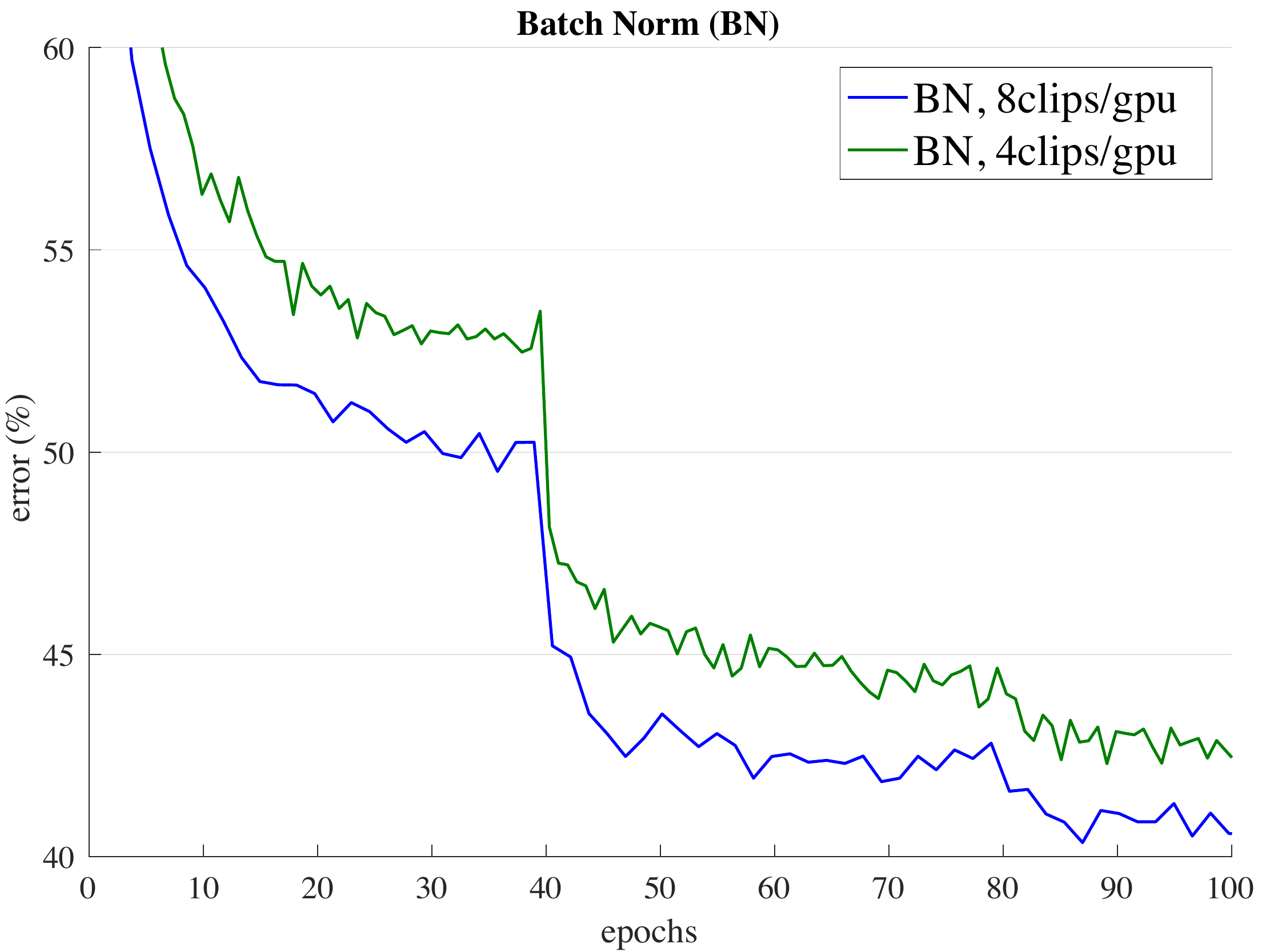}
\includegraphics[width=.4\linewidth]{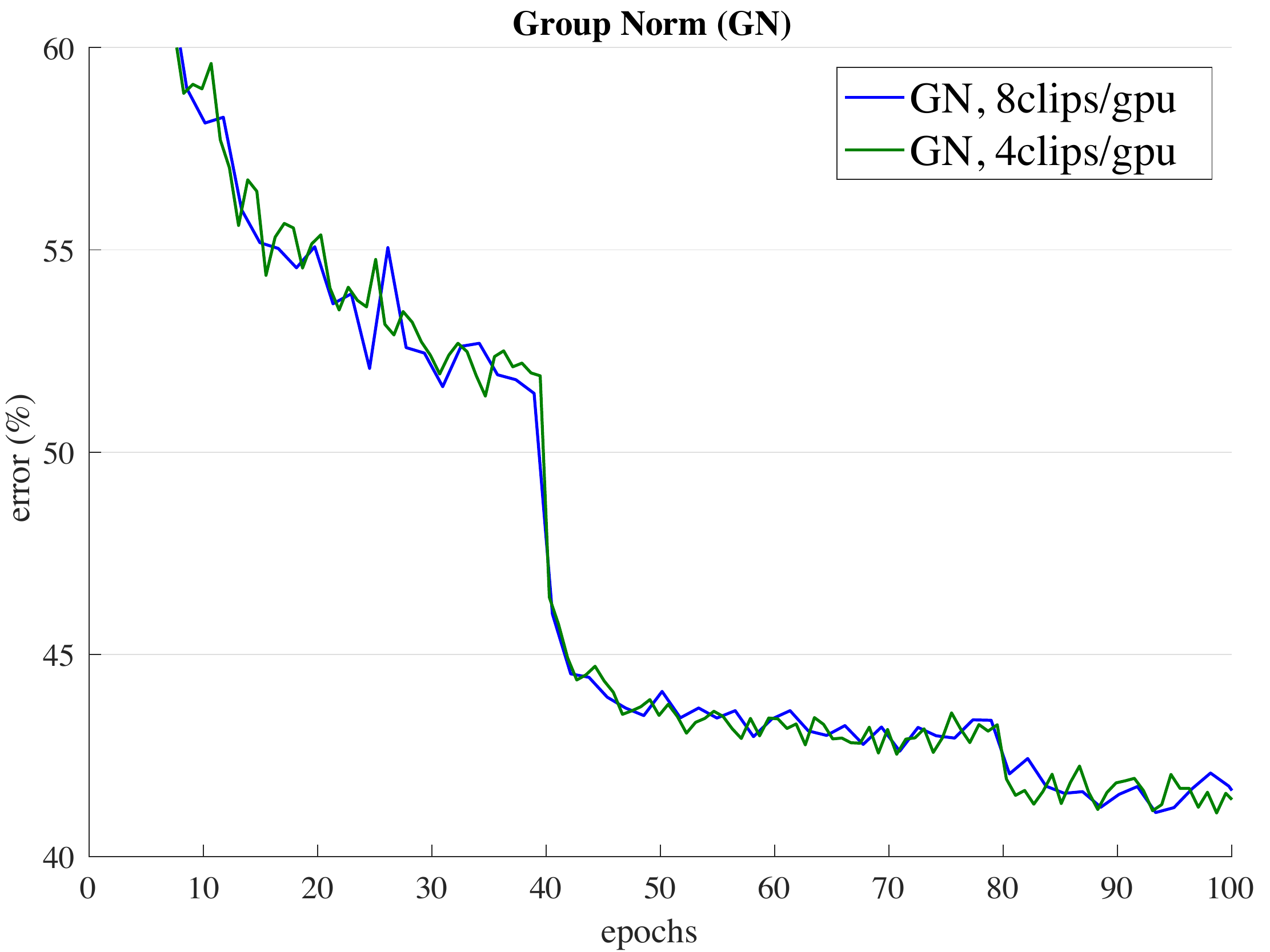}
\vspace{.2em}
\caption{\textbf{Error curves in Kinetics with an input length of 32 frames}. We show ResNet-50 I3D's validation error of BN (left) and GN (right) using a batch size of 8 and 4 clips/GPU. The monitored validation error is the 1-clip error under the same data augmentation as the training set, while the final validation accuracy in Table~\ref{tab:kinetics_bn_gn} is 10-clip testing without data augmentation.}
\label{fig:curves_kinetics}
\vspace{-.3em}
\end{figure*}
%%% 

\paragraph{Results of C4 backbone.} Table~\ref{tab:coco_bn_gn} shows the comparison of GN \vs BN\f on Mask R-CNN using a conv$_4$ backbone (``C4'' \cite{He2017}). This C4 variant uses ResNet's layers of up to conv$_4$ to extract feature maps, and ResNet's conv$_5$ layers as the Region-of-Interest (RoI) heads for classification and regression. As they are inherited from the pre-trained model, the backbone and head both involve normalization layers.

On this baseline, GN improves over BN\f by 1.1 box AP and 0.8 mask AP. We note that the pre-trained GN model is slightly worse than BN in ImageNet (24.1\% \vs 23.6\%), but GN still outperforms BN\f for fine-tuning. BN\f creates inconsistency between pre-training and fine-tuning (frozen), which may explain the degradation.

We have also experimented with the LN variant, and found it is 1.9 box AP worse than GN and 0.8 worse than BN\f. Although LN is also independent of batch sizes, its representational power is weaker than GN.

%%% -----------
\renewcommand\arraystretch{1.15}
\setlength{\tabcolsep}{3.5pt}
\begin{table}[t]
\centering
\small
\begin{tabular}{c|ccc|ccc}
\hline
{\fontsize{7pt}{1em}\selectfont backbone} & AP$^\text{bbox}$ & AP$^\text{bbox}_\text{50}$ & AP$^\text{bbox}_\text{75}$ & AP$^\text{mask}$ & AP$^\text{mask}_\text{50}$ & AP$^\text{mask}_\text{75}$ \\
\shline
BN\f  & 37.7 & 57.9 & 40.9 & 32.8 & 54.3 & 34.7 \\
GN~ & \textbf{38.8} & \textbf{59.2} & \textbf{42.2} & \textbf{33.6} & \textbf{55.9} & \textbf{35.4} \\
\hline
\end{tabular}
\vspace{.5em}
\caption{Detection and segmentation \textbf{ablation results in COCO}, using Mask R-CNN with \textbf{ResNet-50 C4}. BN\f means BN is frozen.
}\label{tab:coco_bn_gn}
\vspace{.3em}
\end{table}
%%% -----------
%%% -----------
\renewcommand\arraystretch{1.15}
\setlength{\tabcolsep}{1.8pt}
\begin{table}[t]
\centering
\small
\begin{tabular}{c|c|ccc|ccc}
\hline
{\fontsize{7pt}{1em}\selectfont backbone} & {\fontsize{7pt}{1em}\selectfont box head} & AP$^\text{bbox}$ & AP$^\text{bbox}_\text{50}$ & AP$^\text{bbox}_\text{75}$ & AP$^\text{mask}$ & AP$^\text{mask}_\text{50}$ & AP$^\text{mask}_\text{75}$ \\
\shline
BN\f  & - & 38.6 & 59.5 & 41.9 & 34.2 & 56.2 & 36.1 \\
BN\f & GN & 39.5 & 60.0 & 43.2 & 34.4 & 56.4 & \textbf{36.3} \\
GN~ & GN & \textbf{40.0} & \textbf{61.0} & \textbf{43.3} & \textbf{34.8} & \textbf{57.3} & \textbf{36.3} \\
\hline
\end{tabular}
\vspace{.5em}
\caption{Detection and segmentation \textbf{ablation results in COCO}, using Mask R-CNN with \textbf{ResNet-50 FPN} and a 4conv1fc bounding box head. BN\f means BN is frozen.
}\label{tab:coco_bn_gn_fpn}
\vspace{.3em}
\end{table}
%%% -----------

%%% -----------
\renewcommand\arraystretch{1.15}
\setlength{\tabcolsep}{2pt}
\begin{table}[t]
\centering
\small
\begin{tabular}{l|ccc|ccc}
\hline
 \multicolumn{1}{c|}{} & AP$^\text{bbox}$ & AP$^\text{bbox}_\text{50}$ & AP$^\text{bbox}_\text{75}$ & AP$^\text{mask}$ & AP$^\text{mask}_\text{50}$ & AP$^\text{mask}_\text{75}$ \\
\shline
R50 BN\f & 38.6 & 59.8 & 42.1 & 34.5 & 56.4 & 36.3 \\
R50 GN & 40.3 & 61.0 & 44.0 & 35.7 & 57.9 & 37.7 \\
R50 GN, long & \textbf{40.8} & \textbf{61.6} & \textbf{44.4} & \textbf{36.1} & \textbf{58.5} & \textbf{38.2}  \\
\hline
R101 BN\f & 40.9 & 61.9 & 44.8 & 36.4 & 58.5 & 38.7 \\
R101 GN & 41.8 & 62.5 & 45.4 & 36.8 & 59.2 & 39.0 \\
R101 GN, long & \textbf{42.3} & \textbf{62.8} & \textbf{46.2} & \textbf{37.2} & \textbf{59.7} & \textbf{39.5} \\
\hline
\end{tabular}
\vspace{.5em}
\caption{\textbf{Detection and segmentation results in COCO} using Mask R-CNN and FPN. Here BN\f is the default Detectron baseline \cite{Detectron2018}, and GN is applied to the backbone, box head, and mask head. ``long'' means training with more iterations. Code of these results are in {\footnotesize \url{https://github.com/facebookresearch/Detectron/blob/master/projects/GN}}.
}\label{tab:coco_bn_gn_final}
\vspace{-.8em}
\end{table}
%%% -----------

\paragraph{Results of FPN backbone.}
Next we compare GN and BN\f on Mask R-CNN using a Feature Pyramid Network (FPN) backbone \cite{Lin2017}, the currently state-of-the-art framework in COCO. Unlike the C4 variant, FPN exploits all pre-trained layers to construct a pyramid, and appends randomly initialized layers as the head. In \cite{Lin2017}, the box head consists of two hidden fully-connected layers (2fc). We find that replacing the 2fc box head with 4conv1fc (similar to \cite{Ren2017b}) can better leverage GN.
The resulting comparisons are in Table~\ref{tab:coco_bn_gn_fpn}.

As a baseline, BN\f has 38.6 box AP using the 4conv1fc head, on par with its 2fc counterpart using the same pre-trained model (38.5 AP). By adding GN to all convolutional layers of the box head (but still using the BN\f backbone), we increase the box AP by 0.9 to 39.5 (2nd row, Table~\ref{tab:coco_bn_gn_fpn}). This ablation shows that a substantial portion of GN's improvement for detection is from \emph{normalization in the head} (which is also done by the C4 variant). On the contrary, applying BN to the box head (that has 512 RoIs per image) does not provide satisfactory result and is $\app$9 AP worse --- in detection, the batch of RoIs are sampled from the same image and their distribution is not \emph{i.i.d.}, and the \emph{non-i.i.d.} distribution is also an issue that degrades BN's batch statistics estimation \cite{Ioffe2017}. GN does not suffer from this problem.

Next we replace the FPN backbone with the GN-based counterpart, \ie, the GN pre-trained model is used during fine-tuning (3rd row, Table~\ref{tab:coco_bn_gn_fpn}).
Applying GN to the backbone \emph{alone} contributes a 0.5 AP gain (from 39.5 to 40.0), suggesting that GN helps when transferring features.

Table~\ref{tab:coco_bn_gn_final} shows the full results of GN (applied to the backbone, box head, and mask head), compared with the standard Detectron baseline \cite{Detectron2018} based on BN\f. Using the same hyper-parameters as \cite{Detectron2018}, GN increases over BN\f by a healthy margin. Moreover, we found that GN is not fully trained with the default schedule in \cite{Detectron2018}, so we also tried increasing the iterations from 180k to 270k (BN\f does not benefit from longer training). 
Our final ResNet-50 GN model (``long'', Table~\ref{tab:coco_bn_gn_final}) is \textbf{2.2} points box AP and \textbf{1.6} points mask AP better than its BN\f variant.

%%% -----------
\renewcommand\arraystretch{1.2}
\setlength{\tabcolsep}{2.2pt}
\begin{table}[t]
\centering
\small
\begin{tabular}{l|ccc|ccc}
\hline
\multicolumn{1}{c|}{\emph{from scratch}} & AP$^\text{bbox}$ & AP$^\text{bbox}_\text{50}$ & AP$^\text{bbox}_\text{75}$ & AP$^\text{mask}$ & AP$^\text{mask}_\text{50}$ & AP$^\text{mask}_\text{75}$ \\
\shline
R50 BN \cite{Li2018} & 34.5 & 55.2 & 37.7 & - & - & - \\
\hline
R50 GN & 39.5 & 59.8 & 43.6 & 35.2 & 56.9 & 37.6 \\
R101 GN & 41.0 & 61.1 & 44.9 & 36.4 & 58.2 & 38.7 \\
\hline
\end{tabular}
\vspace{.5em}
\caption{Detection and segmentation results trained \textbf{from~scratch} in COCO using Mask R-CNN and FPN. Here the BN results are from \cite{Li2018}, and BN is synced across GPUs \cite{Peng2018} and is \emph{not} frozen. Code of these results are in {\footnotesize \url{https://github.com/facebookresearch/Detectron/blob/master/projects/GN}}.
}\label{tab:coco_bn_gn_scratch}
\vspace{-.5em}
\end{table}
%%% -----------

\paragraph{Training Mask R-CNN from scratch.} GN allows us to easily investigate training object detectors \emph{from scratch} (without any pre-training). We show the results in Table~\ref{tab:coco_bn_gn_scratch}, where the GN models are trained for 270k iterations.\footnote{For models trained from scratch, we turn off the default StopGrad in Detectron that freezes the first few layers.}
To our knowledge, our numbers (\textbf{41.0} box AP and \textbf{36.4} mask AP) are the best \emph{from-scratch} results in COCO reported to date; they can even compete with the ImageNet-pretrained results in Table~\ref{tab:coco_bn_gn_final}.
As a reference, with synchronous BN \cite{Peng2018}, a concurrent work \cite{Li2018} achieves a from-scratch result of 34.5 box AP using R50 (Table~\ref{tab:coco_bn_gn_scratch}), and 36.3 using a specialized backbone.

\subsection{Video Classification in Kinetics}

Lastly we evaluate video classification in the Kinetics dataset \cite{Kay2017}. Many video classification models \cite{Tran2015,Carreira2017} extend the features to 3D spatial-temporal dimensions. This is memory-demanding and imposes constraints on the batch sizes and model designs.

We experiment with Inflated 3D (I3D) convolutional networks \cite{Carreira2017}.
We use the ResNet-50 I3D \emph{baseline} as described in \cite{Wang2018}.
The models are pre-trained from ImageNet.
For both BN and GN, we extend the normalization from over $(H, W)$ to over $(T, H, W)$, where $T$ is the temporal axis. We train in the 400-class Kinetics training set and evaluate in the validation set. We report the top-1 and top-5 classification accuracy, using standard 10-clip testing that averages softmax scores from 10 clips regularly sampled.

We study two different temporal lengths: 32-frame and 64-frame input clips. The 32-frame clip is regularly sampled with a frame interval of 2 from the raw video, and the 64-frame clip is sampled continuously. The model is fully convolutional in spacetime, so the 64-frame variant consumes about 2$\times$ more memory. We study a batch size of 8 or 4 clips/GPU for the 32-frame variant, and 4 clips/GPU for the 64-frame variant due to memory limitation.

%%% 
\renewcommand\arraystretch{1.05}
\setlength{\tabcolsep}{8pt}
\begin{table}[t]
\centering
\small
\begin{tabular}{c|c|c|c}
\hline
clip length & 32 & 32 & 64 \\
batch size & 8 & 4 & 4 \\
\shline
BN & \textbf{73.3}~/~\textbf{90.7} & 72.1~/~90.0 & 73.3~/~90.8 \\
GN & 73.0~/~90.6 & \textbf{72.8}~/~\textbf{90.6} & \textbf{74.5}~/~\textbf{91.7} \\
\hline
\end{tabular}
\vspace{1em}
\caption{\textbf{Video classification results in Kinetics}: ResNet-50 I3D baseline's top-1~/~top-5 accuracy (\%).}
\label{tab:kinetics_bn_gn}
\vspace{-1em}
\end{table}

\paragraph{Results of 32-frame inputs.} Table~\ref{tab:kinetics_bn_gn} (col.~1, 2) shows the video classification accuracy in Kinetics using 32-frame clips. For the batch size of 8, GN is slightly worse than BN by 0.3\% top-1 accuracy and 0.1\% top-5. This shows that GN is competitive with BN when BN works well. For the smaller batch size of 4, GN's accuracy is kept similar (72.8~/~90.6 \vs 73.0~/~90.6), but is better than BN's 72.1~/~90.0. BN's accuracy is decreased by 1.2\% when the batch size decreases from 8 to 4.

Figure~\ref{fig:curves_kinetics} shows the error curves. BN's error curves (left) have a noticeable gap when the batch size decreases from 8 to 4, while GN's error curves (right) are very similar.

\paragraph{Results of 64-frame inputs.} Table~\ref{tab:kinetics_bn_gn} (col.~3) shows the results of using 64-frame clips. In this case, BN has a result of 73.3~/~90.8. These appear to be acceptable numbers (\vs 73.3~/~90.7 of 32-frame, batch size 8), but \emph{the trade-off between the temporal length (64 \vs 32) and batch size (4 \vs 8) could have been overlooked}. Comparing col.~3 and col.~2 in Table~\ref{tab:kinetics_bn_gn}, we find that the temporal length actually has positive impact (+1.2\%), but it is veiled by BN's negative effect of the smaller batch size.

GN does not suffer from this trade-off. The 64-frame variant of GN has 74.5~/~91.7 accuracy, showing healthy gains over its BN counterpart and all BN variants. GN helps the model benefit from temporal length, and the longer clip boosts the top-1 accuracy by 1.7\% (top-5 1.1\%) with the same batch size.  

The improvement of GN on detection, segmentation, and video classification demonstrates that GN is a strong alternative to the powerful and currently dominant BN technique in these tasks.

\section{Discussion and Future Work}

We have presented GN as an effective normalization layer without exploiting the batch dimension. We have evaluated GN's behaviors in a variety of applications. We note, however, that BN has been so influential that many state-of-the-art systems and their hyper-parameters have been designed for it, which may not be optimal for GN-based models. It is possible that re-designing the systems or searching new hyper-parameters for GN will give better results.

In addition, we have shown that GN is related to LN and IN, two normalization methods that are particularly successful in training recurrent (RNN/LSTM) or generative (GAN) models. This suggests us to study GN in those areas in the future. We will also investigate GN's performance on learning representations for reinforcement learning (RL) tasks, \eg, \cite{Silver2017}, where BN is playing an important role for training very deep models \cite{He2016}.

\vspace{.5em}
\paragraph{Acknowledgement.} We would like to thank Piotr Doll\'ar and Ross Girshick for helpful discussions.

{
\small
\bibliographystyle{ieee}\bibliography{gn_arxiv.bib}
}

\end{document}